\documentclass[12pt]{article}
\usepackage{arxiv}
\usepackage{amsmath}
\usepackage{graphicx,epstopdf}
\usepackage{times}
\usepackage{graphicx}
\usepackage{color}
\usepackage{multirow}
\usepackage{mathtools}
\usepackage{amssymb}
\usepackage{booktabs}       
\usepackage{amsfonts}       
\usepackage{nicefrac}       
\usepackage[authoryear]{natbib}	
\usepackage{tikz}
\usepackage{algorithm}
\usepackage{algpseudocode}
\usepackage{listings}
\usepackage{newfloat}
\newtheorem{claim}{Claim}

\DeclareFloatingEnvironment[fileext=lop]{Example}

\title{An optimal policy for learning controllable dynamics by exploration}

\date{}

\author{Peter N. Loxley\\
School of Science and Technology\\
University of New England\\
Australia}

\begin{document}
\maketitle

\begin{abstract}

Controllable Markov chains describe the dynamics of sequential decision making tasks and are the central component in optimal control and reinforcement learning. In this work, we give the general form of an optimal policy for learning controllable dynamics in an unknown environment by exploring over a limited time horizon. This policy is simple to implement and efficient to compute, and allows an agent to ``learn by exploring" as it maximizes its information gain in a greedy fashion by selecting controls from a constraint set that changes over time during exploration. We give a simple parameterization for the set of controls, and present an algorithm for finding an optimal policy. The reason for this policy is due to the existence of certain types of states that restrict control of the dynamics; such as transient states, absorbing states, and non-backtracking states. We show why the occurrence of these states makes a non-stationary policy essential for achieving optimal exploration. Six interesting examples of controllable dynamics are treated in detail. Policy optimality is demonstrated using counting arguments, comparing with suboptimal policies, and by making use of a sequential improvement property from dynamic programming. 

\end{abstract} 


\section{Introduction}

Environments with controllable dynamics are interesting to understand from a fundamental viewpoint, as well as providing a potential source of useful applications. When an explicit model of the dynamics is not available one possibility is to learn the dynamics by exploring the environment. This is observed in animal behavior, where an inquisitive animal will often explore a novel environment by actively seeking information about the environment, enabling it to better prepare for future events such as avoiding predators \citep{woodgush,pisula}. Active learning is concerned with a similar task, which involves selecting what data to gather next so as to learn as much as possible about an unknown quantity \citep{mackay1992information}.

\begin{figure}
\centering
\begin{tikzpicture} [scale = 1.5]
\draw[step=0.5cm,color=gray] (0,0) grid (2.5,2.5);
\fill[red] (2,2) rectangle (2.5,2.5) {}; 
\fill[red] (0.5,1.5) rectangle (1,2) {}; 
\fill[red] (1.5,0.5) rectangle (2,1) {}; 
\draw[step=0.5cm,color=black,line width=2pt,cap=round,rounded corners=1pt] 
(0,0) -- (1,0)
(1.5,0) -- (2.5,0)
(2.5,0) -- (2.5,2.5)
(2.5,2.5) -- (1.5,2.5)
(1,2.5) -- (0,2.5)
(0,2.5) -- (0,0)
(1.5,2) -- (1.5,2.5)
(1.5,2) -- (2,2)
(0,2) -- (1,2)
(1,2) -- (1,1.5)
(1,1.5) -- (2,1.5)
(0.5,1) -- (0.5,1.5)
(0,1) -- (0.5,1)
(1,1) -- (1.5,1)
(1,0.5) -- (1,1)
(2,0.5) -- (2,1)
(0.5,0.5) -- (2,0.5)
(1.5,0) -- (1.5,0.5);
\end{tikzpicture}
\caption{A simple environment given by a $5\times 5$ maze with three trapping states (red squares). An agent can explore this environment to learn the maze but will likely encounter a trapping state and become trapped along the way. When this happens the agent can no longer continue to explore.}
\label{fintro}
\end{figure}
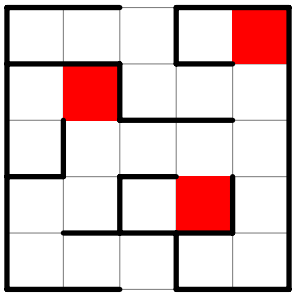

In this work, we are interested in the structure of optimal policies for exploring unknown environments to learn controllable dynamics. This is a difficult problem in general, as illustrated by the following example. Consider the simple environment in Figure \ref{fintro} given by a $5\times 5$ maze with three trapping states (in Markov chains these are called absorbing states). It is certainly possible for an agent to explore this environment and learn the structure of the maze in the process of doing so. The dilemma the agent faces is where to explore? Exploring every part of the environment means the agent will eventually encounter a trapping state and become trapped; rendering it unable to continue exploring, and most likely leading to an incomplete knowledge of the maze. Other types of states can also interfere with exploration, as we shall see. It is therefore desirable for an agent to avoid certain states at certain times during exploration. It is also desirable for an agent to explore new locations that have not yet been explored in order to learn new information. Ideally, a policy for exploring would aim to strike an optimal balance between these somewhat opposing objectives.  

To address this challenge we apply the widely used framework of controllable Markov chains to describe the dynamics of an unknown environment. We then investigate exploration and learning within such an environment. This is done by making use of information theory and optimal control, building on the idea of optimal experimental design pioneered by Kristen Smith \citep{smith1918standard}, and further developed in works since \citep{lindley1956measure,pfaffelhuber, mackay1992information, oaksford1994rational,storck1995reinforcement,little,loxley}. More specifically, we seek the form of an optimal policy for an agent to learn the transition probabilities of a controllable Markov chain by exploring an unknown environment using a limited number of exploration steps (i.e., a limited time horizon). The existence of certain types of states that restrict control of the dynamics (including transient states, absorbing states, and non-backtracking states) makes planning essential for exploration to be optimal. 

Early related work in reinforcement learning includes the model-free approaches given by Q-Learning \citep{watkins}, and the Dyna-Q algorithm \citep{sutton}. The Dyna-Q algorithm was specifically designed for solving an optimal control task while simultaneously learning a model of the environment to enable planning (i.e., to speed up policy convergence). However, the learned model of the environment is a by-product of solving a generic optimal control task, rather than the specific task of optimal exploration investigated here. Presumably, it would be possible to apply Dyna-Q to this specific task. However, Dyna-Q makes use of Q-learning (or more generally, some infinite horizon learning algorithm), which has certain limitations that are discussed next. 

More closely-related work appears in \cite{storck1995reinforcement}, where Q-learning was applied directly to the task of optimal exploration. However, Q-learning is an infinite horizon technique leading to stationary policies, and so is not able to capture the necessary temporal structure and non-stationary policies required for optimal exploration over a finite time horizon. This can be seen in the Q-learning algorithm, where learning episodes must be repeated many times over to ensure convergence (the proof of convergence for Q-Learning requires that every state and control pair is visited infinitely often \citep{watkinsdyan,tsi}). We will discuss this point further in the Discussion. Clearly an alternative approach must be considered for optimal exploration over finite time horizons.

The most closely related previous work is in \cite{little}, and \cite{loxley2025}. In \cite{little}, the predicted information gain was proposed as a useful information measure for exploration, and  exploration strategies maximizing this measure approximately were investigated. Connections between exploration and curiosity in humans and animals were also discussed. In \cite{loxley2025}, finite-horizon optimal control techniques were applied to exploration. In both works, the influence of absorbing states on exploration was recognized and investigated. The contribution given by the present work greatly extends and generalizes some of these findings. 

In Sec. \ref{theory}, our approach and supporting theory for optimal exploration is outlined, and the general form of an optimal policy is proposed. In Sec. \ref{results}, we present a control set parameterization, before describing an algorithm for finding an optimal policy, and applying it to treat six examples of controllable dynamics in detail. A brief discussion of information measures is also provided. In Sec. \ref{conc}, we present a conclusion and discussion.

\section{Informative Exploration}\label{theory}

An agent or controller explores an unknown environment sequentially over a finite number of time periods by learning how the environment responds after choosing a particular control from a given state. Assuming a finite set of states $S$ and a finite set of controls $U$, choosing control $u\in U$ moves an environment from state $i$ to state $j$ according to a set of unknown transition probabilities $p_{ij}(u)$ of a controllable Markov chain (CMC). By following the dynamics of a CMC an agent forms estimates of the unknown transition probabilities. An informative policy tells the agent where to explore next to gain the most information about these probabilities, leading to efficient exploration.

It is possible to estimate transition probabilities for some underlying controllable dynamics by counting transitions between different states while exploring the environment. These counts are stored in a tensor $\boldsymbol{F}$; where tensor component $F_{uij}$ is the number of transitions from state $i$ to state $j$ when control $u$ is chosen. A reasonable estimate is then given by \mbox{$\widehat{p}_{ij}(u,\boldsymbol{F})=(F_{uij}+\alpha)/\sum_{j'}(F_{uij'}+\alpha)$}. This estimate corresponds to the mean of a Dirichlet distribution with parameters $\boldsymbol{F}$ and $\alpha$. In the present case, this Dirichlet distribution represents the posterior probability of each distribution $p_{i\boldsymbol\cdot}(u)$, given the data $\boldsymbol{F}$, and a Dirichlet prior with parameter $\alpha$ \citep{mackay}.

To arrive at a good CMC estimate requires some form of informative exploration, visiting key states and transitions that have not previously been visited. In this work, informative exploration means using an information measure $h(i,u,\boldsymbol{F})$ to evaluate the potential of  control $u$ for exploring new states accessible from state $i$, depending on the past exploration history given in $\boldsymbol{F}$. A large value of $h(i,u,\boldsymbol{F})$ implies a large information gain about the unknown probability distribution $p_{i\boldsymbol\cdot}(u)$. The simplest approach to informative exploration is given by a \emph{greedy policy} that chooses control $u$ in state $i$ according to the rule:
\begin{align}
&\text{maximize}\ h(i,u,\boldsymbol{F})\label{greedy}, \\
&\text{subject to}\ u\in U.\nonumber
\end{align}
This greedy choice for $u$ maximizes the current value of $h$ without regard to any future values of $h$, and generally leads to suboptimal policies. The reason is that it may be possible to obtain larger values of $h$ at future time periods by choosing controls that do not necessarily maximize $h$ at earlier time periods. 

An \emph{optimal policy} requires maximizing the sum of $h$ over $N$ time periods of exploration:
\begin{equation}
J_\pi(i_0,\boldsymbol{F}_0)={\mathbb{E}}\left \{\sum_{k=0}^{N-1}h(i_k,\mu_k,\boldsymbol{F}_k)\right\},\label{totalcost}
\end{equation}
where the expectation is taken over the joint distribution of the states $i_1,...,i_{N-1}$, and can be found from the ``initial state" $(i_0,\boldsymbol{F}_0)$ and the transition probabilities $p_{ij}(\mu_k(i,\boldsymbol{F}))$; where policy $\pi=\{\mu_0,...,\mu_{N-1}\}$ is a sequence of functions $\mu_k$ that map states into controls: $\mu_k(i,\boldsymbol{F})=u$. A general starting point for finding an optimal policy is provided by the framework of exact dynamic programming \citep{bert3}. In this case the finite-horizon dynamic programming algorithm takes the form:
\begin{equation}
J_k^*(i,\boldsymbol{F}) = \underset{u\in U_k(i)}{\operatorname{max}}\ \Big [h(i,u,\boldsymbol{F})+\sum_j p_{ij}(u)J_{k+1}^*(j,F_{uij}+1)\Big ],\label{dp}
\end{equation}
where $F_{uij}$ is updated to $F_{uij}+1$ with probability $p_{ij}(u)$, corresponding to a transition from state $i$ to state $j$ under control $u$. This equation is iterated backwards from time period $k=N-1$ to time period $k=0$, for each possible state, yielding the optimal sum-to-go $J_{k}^*$, and optimal policy $\mu^*$, as
\begin{equation}
\mu_k^*(i,\boldsymbol{F}) = \underset{u\in U_k(i)}{\operatorname{arg\ max}}\ \Big [h(i,u,\boldsymbol{F})+\sum_j p_{ij}(u)J_{k+1}^*(j,F_{uij}+1)\Big ].\label{optpol}
\end{equation}
Unfortunately, backward iteration of (\ref{dp}) is not possible for two reasons. The first is that we do not know $p_{ij}(u)$, although we assume it is possible to draw samples $j\sim p_{i\boldsymbol\cdot}(u)$ using simulation. The second is that (\ref{dp}) suffers from Bellman's curse of dimensionality. The ``state" is now given by the pair $(i,\boldsymbol{F})$, and since $\boldsymbol{F}$ depends on the entire history of exploration, the size of the corresponding state-space grows exponentially over time. 

A more natural approach to exploring an environment is to apply simulation; visiting states and exploring different controls over a number of time periods. In this case, simulation can be used to approximate the optimal sum-to-go $J_{k+1}^*$ as $\tilde{J}_{k+1}$, yielding the \emph{one-step lookahead approximation} of (\ref{optpol}) as
\begin{equation}
\bar{\mu}_k(i,\boldsymbol{F}) = \underset{u\in U_k(i)}{\operatorname{arg\ max}}\ \Big [h(i,u,\boldsymbol{F})+\sum_j p_{ij}(u)\tilde{J}_{k+1}(j,F_{uij}+1)\Big ].\label{onestep}
\end{equation}
The policy $\bar{\mu}$ will be a close approximation of the optimal policy $\mu^*$ provided $\tilde{J}$ is a close approximation of $J^*$. The one-step lookahead approximation given by (\ref{onestep}) forms the basis for a popular class of methods in approximate dynamic programming and reinforcement learning.  

We are now in a position to state the central claim of this work. By restricting the class of policies to a simple parametric form, we can generalize the greedy rule given in (\ref{greedy}) to
\begin{equation}
\mu_k(i,\boldsymbol{F},r) = \underset{u\in U_k(i,r)}{\operatorname{arg\ max}}\  h(i,u,\boldsymbol{F}),\label{parametric}
\end{equation}
where $\mu$ is a parametric policy, and $U_k(i,r)$ is a parametric set of controls with parameter $r$. Once a parametric form for $U_k(i,r)$ has been chosen, the value of $r$ can be found by searching directly in the space of policies. Given an initial state $(i_0,\boldsymbol{F}_0)$ and a policy $\mu_k(i,\boldsymbol{F},r)$, the objective given by (\ref{totalcost}) depends only on $r$ and the unknown transition probabilities. Applying simulation (and possibly approximation), it then becomes possible to determine $r$ through the maximization:
\begin{align}
&\text{maximize}\ J_r(i_0,\boldsymbol{F}_0)\label{jmax},\\
&\text{subject to}\ r\in R.\nonumber
\end{align}
This leads directly to the following claim:
\begin{claim} 
A simple parametric choice exists for $U_k(i,r)$ that leads to $\mu_k(i,\boldsymbol{F},r^*)$ as an optimal (or near-optimal) policy for learning CMCs by informative exploration.
\end{claim}

The content of this claim is that optimal policies for informative exploration all have the same general structure given in (\ref{parametric}): greedy maximization of an information measure over a set of controls that depends on both time and state. For this reason, the parametric policy given by (\ref{parametric}) could also be called an \emph{informative policy}. More specifically, a control set that changes over time is able to capture the key difficulties of exploring non-trivial environments using planning and lookahead over many time periods. Furthermore, the parametric form for $U_k(i,r)$ turns out to be simple to understand and interpret. This claim will now be verified for a collection of interesting examples.

\section{Results}\label{results}

\subsection{Control set parameterization}

Exploration is a challenging problem due the existence of states that restrict control of the dynamics. Restrictive states include absorbing states and non-backtracking states; but more generally, any type of state that gives rise to one or more transient states. The key property of a restrictive state is that it limits the opportunity for further exploration. The presence of restrictive states in a CMC usually requires some form of planning or lookahead in order for exploration to be successful. The CMCs investigated here all have one or more restrictive states. Motivated by the simulated policies in \cite{loxley2025}, we now propose a parametric form for the control set in (\ref{parametric}).

Defining $S'(r)$ to be the set of states leading directly to a restrictive state (excluding the restrictive state itself): $S'(r)=\{i\ |\ \exists u,\ p_{ij}(u)>0 \text{ and } j =\text{``restrictive state"} \text{ and } i \neq j\}$, and defining $U'(i,r)$ to be the set of controls leading directly to a restrictive state from one of these states: $U'(i,r)=\{u\ |\ p_{ij}(u)>0 \text{ and } j =\text{``restrictive state"} \text{ and } i \neq j\}$, the parametric form for $U_k(i,r)$ is given by:
\begin{equation}
U_k(i,r) = 
\begin{cases}
U_k(i) - U'(i,r)&\text{if } i \in S'(r) \text{ and } k < t_i,\\
U_k(i) & \text{otherwise}.\label{par1}
\end{cases}
\end{equation}
where $t_i$ is some time constant. The set-difference in (\ref{par1}) prevents any controls being applied that would lead the current state directly to a restrictive state before some time period $t_i$ has passed. After time period $t_i$, any control can be applied, regardless of the state of the environment. This parameterization prevents the environment from entering a restrictive state prematurely, thereby allowing exploration of (transient) states and controls that would otherwise be inaccessible. 

\subsection{Algorithm}

Searching in policy space requires evaluating the objective (\ref{totalcost}) by simulating a parametric policy for a given value of $r$ and an initial state $(i_0,\boldsymbol{F}_0)$. This can be performed as follows: At time period $k$ and state $(i,\boldsymbol{F})$, we apply control $\mu_k(i,\boldsymbol{F},r)=u$ using our parametric policy, then draw sample $j\sim p_{i\boldsymbol\cdot}(u)$ using simulation, before undergoing a state transition to $(j,F_{uij}+1)$ and arriving at time period $k+1$. At each time period we accumulate the value of $h(i_k,u_k,\boldsymbol{F}_k)$ to form the sum in (\ref{totalcost}). Simulating $N$ time periods in this way (starting at $k=0$ and finishing at $k=N-1$) yields a single trajectory. To compute the expectation in (\ref{totalcost}), we simulate a ``large" number of these trajectories and take a Monte Carlo average. This yields the value of $J_r(i_0,\boldsymbol{F}_0)$ required in (\ref{jmax}).

For simple environments describable by small CMCs, optimization in policy space can be performed by exhaustive search; i.e., by considering all possible discrete values of $r\in R$ in the maximization in (\ref{jmax}). Exploration becomes more difficult in complex environments with large CMCs, where parameterized policies now occupy larger parameter spaces. The cross-entropy method (CEM) is a smart random search technique that has been extensively used to learn controllers in the game of tetris \citep{szita,thiery}. Here, we use CEM for optimization in policy space when exhaustive search becomes intractable. In this work, CEM is implemented using the binomial distribution. We start by using a binomial distribution with parameters $n_i-1$ and $p_i$ to generate many candidate values for the control set parameter $r_i-1$ (since the binomial distribution starts at 0 we subtract 1 from $n_i$ and $r_i$ for the purpose of sample generation, before adding 1 back afterwards). Doing this for each control set parameter, then selecting a small percentage of the top candidates according to (\ref{totalcost}), allows each binomial distribution to be updated by using these top candidates to estimate new values for each $p_i$ (each $n_i$ is determined by the fixed range of each $r_i$). These new values are given by $p_i=\bar{x}_i/n_i$, where $\bar{x}_i$ is the mean of the top candidates for $r_i$. The ``generate", ``select", and ``update" steps are repeated a number of times until the binomial distributions converge to a distribution of (locally) optimal parameter values. Unlike using the Gaussian distribution in CEM, it is unnecessary to regenerate parameter distributions by injecting additional noise after each ``select" step. 

Approximation in value space (\ref{onestep}) is now used on top of approximation in policy space to look for improvements in a suboptimal policy (if the policy is already optimal then no improvements are possible). More specifically, the parametric policy in (\ref{parametric}) with its optimal $r$ value (denoted $r^*$) is used as a base policy in the rollout algorithm  \citep{tesauro,bert_roll} for determining (\ref{onestep}). The rollout algorithm works as follows: At time period $k$ and state $(i,\boldsymbol{F})$, we choose control $u$, then draw sample $j\sim p_{i\boldsymbol\cdot}(u)$ using simulation, before undergoing a state transition to $(j,F_{uij}+1)$ and arriving at time period $k+1$. From here, we use the base policy to apply control $\mu_{k+1}(i_{k+1},\boldsymbol{F}_{k+1},r^*)$ at time period $k+1$. The simulation continues with the base policy determining the control at each time period, and each value of $h$ is included into a sum that approximates $\tilde{J}_{k+1}$ in (\ref{onestep}), until reaching the horizon when $k=N-1$. In the case of stochastic dynamics, a large number of trajectories from $k$ to $N-1$ are simulated, and Monte Carlo averaging is used to arrive at an approximate value for the sum on the right-hand-side of (\ref{onestep}). This process is repeated for each control $u\in U_k(i)$, and the one-step lookahead in (\ref{onestep}) is used to obtain the rollout policy $\bar{\mu}_k$. In this way, the base policy $\mu_k$ has been ``improved" at time period $k$ by the rollout policy $\bar{\mu}_k$. 

A sequential improvement property \citep{bert3}, guarantees the rollout policy $\bar{\pi}$ is no worse than the base policy $\pi$: $J_{\bar{\pi}}(i_0,\boldsymbol{F}_0)\geq J_{\pi}(i_0,\boldsymbol{F}_0)$, meaning that $\bar{\mu}_k(i,\boldsymbol{F})$ is always at least as good as $\mu_k(i,\boldsymbol{F},r^*)$. A consequence of this fact is that when $J_{\pi}(i_0,\boldsymbol{F}_0)$ equals $J_{\bar{\pi}}(i_0,\boldsymbol{F}_0)$, the parametric policy cannot be improved by one-step lookahead with rollout. This is true when the parametric policy is an optimal policy, but the converse is not always true.

\subsection{Example 1}

The first example we consider is the CMC with two states and two controls shown in Example \ref{e1}. For control one, the Markov chain has a transient state (state one) and an absorbing state (state two) provided $p<1$; while for control two, each state is a recurrent state. 
\begin{Example}
\centering
\begin{tikzpicture}[->,shorten >=2pt,line width=0.5pt,node distance=2cm]
\node [circle,draw] (one) {1};
\node [circle,draw] (two) [right of=one] {2};
\path (one) edge [loop left] node [left] {$p$} (one);
\path (one) edge [line to] node [above] {$1-p$} (two);
\path (two) edge [loop right] node [right] {$1$} (two);
\node at (1,1.4) {\color{blue}{$u=1$}};
\node [circle,draw] (one-u2) [below of=one] {1};
\path (one-u2) edge [loop left] node [left] {$1$} (one-u2);
\node [circle,draw] (two-u2) [below of=two] {2};
\path (two-u2) edge [loop right] node [right] {$1$} (two-u2);
\node at (1,-1.2) {\color{blue}{$u=2$}};
\end{tikzpicture}
\caption{A controllable Markov chain (CMC) with two states and two controls.}
\label{e1}
\end{Example}
The absorbing state makes it difficult to estimate the transition probabilities of this CMC. Effective exploration requires choosing a sequence of controls that allows state one to be adequately explored before reaching the absorbing state. 

Assume we know nothing about the transition probabilities except that one of the states leads to an absorbing state under one of the controls. We do not know which state, or which control, this could be; so we would like to learn the sets $S'(r)$ and $U'(i,r)$ in (\ref{par1}), as well as the time constant $t_i$. Making use of our assumptions, we require three parameters: $r=(r_1,r_2,r_3)$; where $r_1\in\{1,2\}$ is the unknown state, $r_2\in\{1,2\}$ is the unknown control, and $r_3\in\{1,...,N\}$ is the unknown time constant.

\begin{figure}
\centering
\includegraphics[scale=0.8,bb=0 320 580 520,clip=true]{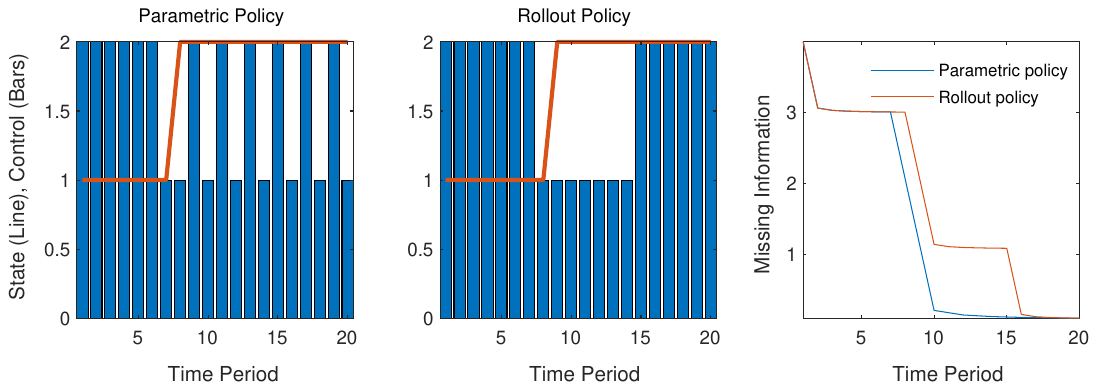}
\caption{Optimal policies for Example \ref{e1}. (Left Panel) Applying the parametric policy to the CMC in Example \ref{e1} for twenty time periods leads to the states (line) and controls (bars) shown for each time period. For example, during time period 1 the CMC is in state 1 and control 2 has been selected by the policy. (Center Panel) States and controls for the rollout policy. (Right Panel) The corresponding decrease in missing information at each time period is shown for each policy.}
\label{f1}
\end{figure}

Figure \ref{f1} shows two state trajectories for the CMC in Example \ref{e1}. For each trajectory, the environment begins in state one, and dynamically evolves under one of the policies for twenty time periods.
The dynamics is completely deterministic ($p=0$) in this case. The parametric policy can be optimized using exhaustive search and the optimal parameter values are found to be $r=(1,1,7)$; correctly identifying the state and control leading to the absorbing state. 

In Figure \ref{f1} (Left Panel), the parametric policy samples from the transition probability $p_{11}(2)$ for six time periods. During the seventh time period the policy selects control one, which has just become available in control set, and the environment moves to state two: so that $p_{12}(1)$ is sampled once. Following this, $p_{22}(1)$ and $p_{22}(2)$ are sampled alternately over thirteen time periods. This policy leads to an almost equal division of ninteen sampling periods over three transition probabilities: giving 6, 7, and 6. Therefore, in this example the parametric policy is an optimal policy for exploration. In Figure \ref{f1} (Center Panel), the rollout policy is also an optimal policy for exploration, as expected from the sequential improvement property, and leads to a slightly different division of sampling periods as 7, 6, 6. 

In Figure \ref{f1} (Right Panel), the decrease in missing information is shown at each time period for both the parametric and rollout policies. The missing information is simply the KL divergence between the true distribution $p_{i\cdot}(u)$, and the estimated distribution $\widehat{p}_{i\cdot}(u,\boldsymbol{F})$, summed over all $i$ and $u$ (and with $\alpha$ set to 0.05 to reflect our prior model ignorance). For both policies, the missing information decreases from $4$ to $0.103$ over twenty time periods; with the first decrease due to learning the estimate for $p_{11}(2)$, and the second decrease (or second and third decreases) due to learning estimates for $p_{22}(1)$ and $p_{22}(2)$. Notice that the missing information initially decreases very rapidly when learning a new estimate, followed by a more gradual decrease as that estimate is refined over time.

\begin{figure}
\centering
\includegraphics[scale=0.8,bb=0 320 580 520,clip=true]{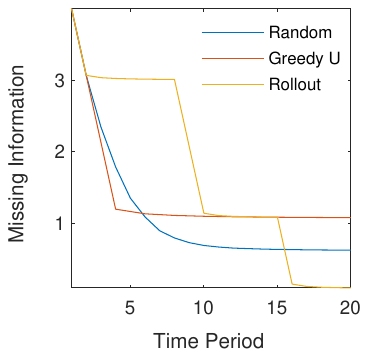}
\caption{Suboptimal policies for Example \ref{e1}. The decrease in missing information at each time period is shown for the uniform random policy over unrestricted controls (Random), and the greedy policy over unrestricted controls (Greedy U). The rollout policy with a Greedy U base policy is shown for comparison (Rollout).}
\label{f2}
\end{figure}

A comparison with two suboptimal policies is shown in Figure \ref{f2}. 
Greedy maximization of the information measure over unrestricted controls leads to a division of sampling periods as either 0, 9, 10 or 1, 9, 9. In the first case, control one is processed first, so the environment immediately moves to the absorbing state before the policy alternates between the two available controls. In the second case, control two is processed first, so the environment remains in state one for a single time period, before control one is applied during the second time period and the environment moves to the absorbing state. Figure \ref{f2} shows the case where control one is processed first. A uniform random policy over unrestricted controls leads to a division of sampling periods as 0, 9, 10 (with probability $1/2$) and 1, 9, 9 (with probability $1/2$). Figure \ref{f2} shows a Monte Carlo average of these two cases. Each of these policies clearly leads to suboptimal exploration. Figure \ref{f2} also shows the rollout policy for a base policy given by the greedy policy over unrestricted controls. In this case, the rollout policy is able to significantly improve its base policy and generates the optimal policy shown in Figure \ref{f1}. 

\begin{figure}
\centering
\includegraphics[scale=0.8,bb=0 320 580 520,clip=true]{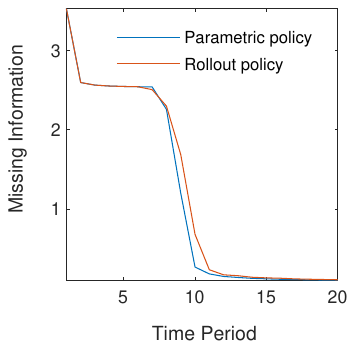}
\caption{Policies for Example \ref{e1} with stochastic dynamics ($p=0.1$). The decrease in the Monte Carlo averaged missing information is shown for parametric and rollout policies.}
\label{f3}
\end{figure}

State trajectories obeying stochastic dynamics (with $p=0.1$) are shown in Figure \ref{f3} for parametric and rollout policies. In this case, Monte Carlo averaging (over 100 trajectories) leads both trajectories to look very similar (c.f. Figure \ref{f1}). We treat stochastic dynamics in fuller detail in Example 6.

\subsection{Example 2}

In this example we consider a deterministic CMC with four states and four controls, as shown in Example \ref{e2}. 
\begin{Example}
\centering
\begin{tikzpicture}[->,shorten >=2pt,line width=0.5pt,node distance=2cm]
\node [circle,draw] (one) {1};
\path (one) edge [line to] node [above] {$1$} (two);
\node [circle,draw] (two) [right of=one] {2};
\path (two) edge [loop right] node [right] {$1$} (two);
\node [circle,draw] (three) [right of=two] {3};
\path (three) edge [loop right] node [right] {$1$} (three);
\node [circle,draw] (four) [right of=three] {4};
\path (four) edge [loop right] node [right] {$1$} (four);
\node at (3,1.0) {\color{blue}{$u=1$}};

\node [circle,draw] (one-2) [below of=one] {1};
\path (one-2) edge [loop left] node [left] {$1$} (one-2);
\node [circle,draw] (two-2) [right of=one-2] {2};
\node [circle,draw] (three-2) [right of=two-2] {3};
\path (two-2) edge [line to] node [above] {$1$} (three-2);
\path (three-2) edge [loop right] node [right] {$1$} (three-2);
\node [circle,draw] (four-2) [right of=three-2] {4};
\path (four-2) edge [loop right] node [right] {$1$} (four-2);
\node at (3,-1.0) {\color{blue}{$u=2$}};

\node [circle,draw] (one-3) [below of=one-2] {1};
\path (one-3) edge [loop left] node [left] {$1$} (one-3);
\node [circle,draw] (two-3) [right of=one-3] {2};
\path (two-3) edge [loop right] node [right] {$1$} (two-3);
\node [circle,draw] (three-3) [right of=two-3] {3};
\node [circle,draw] (four-3) [right of=three-3] {4};
\path (three-3) edge [line to] node [above] {$1$} (four-3);
\path (four-3) edge [loop right] node [right] {$1$} (four-3);
\node at (3,-3.0) {\color{blue}{$u=3$}};

\node [circle,draw] (one-4) [below of=one-3] {1};
\path (one-4) edge [loop left] node [left] {$1$} (one-4);
\node [circle,draw] (two-4) [right of=one-4] {2};
\path (two-4) edge [loop right] node [right] {$1$} (two-4);
\node [circle,draw] (three-4) [right of=two-4] {3};
\path (three-4) edge [loop right] node [right] {$1$} (three-4);
\node [circle,draw] (four-4) [right of=three-4] {4};
\path (four-4) edge [loop right] node [right] {$1$} (four-4);
\node at (3,-5.0) {\color{blue}{$u=4$}};
\end{tikzpicture}
\caption{A deterministic CMC with four states and four controls.}
\label{e2}
\end{Example}
For the first three controls, each Markov chain has a transient state, an absorbing state, and two recurrent (non-absorbing) states. For the fourth control there are no absorbing states, all states are recurrent. Accurately estimating the transition probabilities by exploring the environment described by this CMC is clearly going to be challenging. 

Once again, assume we know nothing about the transition probabilities except that there are three unknown states and three unknown controls that lead to an absorbing state over different unknown durations. To learn the sets $S'(r)$ and $U'(i,r)$ in (\ref{par1}), as well as the time constants $t_i$, we make use of these assumptions to propose $r=(r_1,...,r_9)$; where the first three parameters correspond to unknown states, the next three parameters correspond to unknown controls for each of these states, and the last three parameters correspond to unknown time constants for each of these states. 

\begin{figure}
\centering
\includegraphics[scale=0.8,bb=0 320 580 520,clip=true]{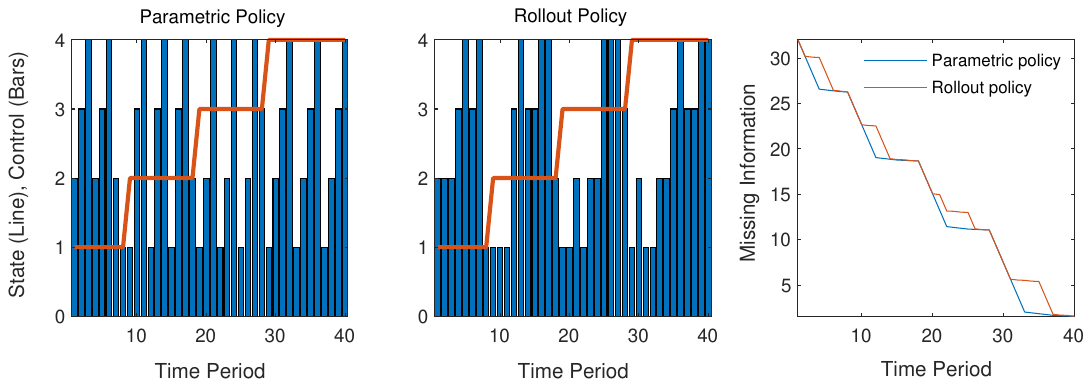}
\caption{Optimal policies for Example \ref{e2}. (Left Panel) Parametric policy applied to the CMC in Example \ref{e2}. (Center Panel) Rollout policy. (Right Panel) Decrease in missing information for each policy.}
\label{f4}
\end{figure}

Figure \ref{f4} shows two state trajectories for the CMC in Example \ref{e2}. For each trajectory, the environment begins in state one, and dynamically evolves under one of the policies for forty time periods. 
In this case, the parametric policy is optimized using CEM as the parameter space is now too large for exhaustive search. The optimal parameter values are found to be $r=(1,2,3,1,2,3,8,18,28)$; correctly identifying each state-control pair leading to an absorbing state. 

In Figure \ref{f4} (Left Panel), the parametric policy samples from the transition probabilities $p_{11}(2),p_{11}(3),$ and $p_{11}(4)$ for seven time periods. During the eighth time period the policy selects control one, which has just become available in control set, and the environment moves to state two: so that $p_{12}(1)$ is sampled once. Following this; $p_{22}(1),p_{22}(3),$ and $p_{22}(4)$ are sampled alternately over nine time periods; $p_{33}(1),p_{33}(2),$ and $p_{33}(4)$ are sampled alternately over nine time periods; and $p_{44}(1),...,p_{44}(4)$ are sampled alternately over twelve time periods. This is the most even division possible of 37 sampling periods into 13 transition probabilities (i.e., two of the transition probabilities must be sampled twice instead of three times). The remaining three transition probabilities only receive one sampling period each: this is the maximum number possible for a transition departing from a transient state. Therefore, in this example the parametric policy is an optimal policy for exploration. 

In Figure \ref{f4} (Center Panel), the rollout policy leads to the same division of sampling periods as the parametric policy does, although the controls are applied in a different order. In Figure \ref{f4} (Right Panel), the decrease in missing information is shown at each time period for both the parametric and rollout policies. For both policies, the missing information decreases from $32$ to $1.58$ over forty time periods.

In the absence of planning or lookahead, the worst-case division of 37 sampling periods would be 0, 0, 0, 37 when control one, two, and three are chosen in order. The best-case scenarios for the greedy policy over unrestricted controls, and the uniform random policy, would be a division of sampling periods given by 3, 3, 3, 28. The greedy policy alternates between all four controls at each state, while the uniform random policy applies each control with probability $1/4$. The geometric distribution with parameter $1/4$ has a mean of 4, so an absorbing-state control would be selected by the uniform random policy after at most 3 trials, on average. Each of these policies clearly leads to suboptimal exploration. 

\subsection{Example 3}

Our next example is a $4\times 4$ gridworld with a single absorbing state, as shown in Example \ref{e3}. States are labelled from one to sixteen across the four rows of the grid, and the absorbing state is located at state sixteen in the bottom right-hand corner. 
\begin{Example}
\centering
\begin{tikzpicture} [scale = 1.5]
\draw[step=0.5cm,color=gray] (0,0) grid (2,2);
\draw[step=0.5cm,color=black,line width=1.2pt,cap=round,rounded corners=1pt] 
(0,0) -- (2,0)
(2,0) -- (2,2)
(2,2) -- (0,2)
(0,2) -- (0,0);
\fill[red] (1.5,0) rectangle (2,0.5) {}; 
\end{tikzpicture}
\caption{A $4\times 4$ gridworld with one absorbing state (red square). Each CMC transition probability is one for a move of one square along the cardinal direction specified by a control (i.e., `up', `down', `left', or `right'), and zero otherwise; unless that move would cross the boundary; in which case, the agent will remain in the same square with probability one.}
\label{e3}
\end{Example}
There are four possible controls available at each state; with each control corresponding to a move of one square along one of the cardinal directions: $1$ = `up', $2$ = `down', $3$ = `left', $4$ = `right'. Each CMC transition probability is one for a move of one square along the cardinal direction specified by a control, and zero otherwise; unless that move would cross the boundary; in which case, the agent will remain in the same square with probability one.

Usual tasks on gridworlds require solving problems such as finding the shortest path from a starting location to the absorbing state. However, informative exploration is closer to a \emph{longest-path} problem, where the objective is to explore all states before reaching the absorbing state. 

Now assume we know nothing about the transition probabilities except that there is a single absorbing state in one of the corners of the gridworld. To learn the sets $S'(r)$ and $U'(i,r)$ in (\ref{par1}), as well as the time constants $t_i$, we make use of this assumption to propose $r=(r_1,...,r_5)$; where the first two parameters correspond to two unknown states, the next two parameters correspond to one unknown control for each of these states, while the last parameter corresponds to a single unknown time constant applied to both of these states. 

\begin{figure}
\centering
\includegraphics[scale=0.8,bb=0 320 580 520,clip=true]{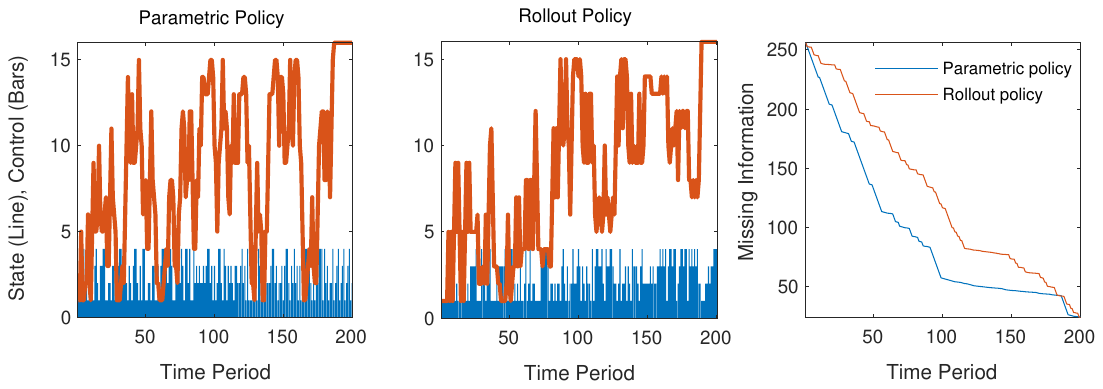}
\caption{Near-optimal policies for Example \ref{e3}. (Left Panel) Parametric policy applied to the CMC corresponding to the gridworld in Example \ref{e3}. (Center Panel) Rollout policy. (Right Panel) Decrease in missing information for each policy.}
\label{f5}
\end{figure}

Figure \ref{f5} shows two state trajectories for the CMC corresponding to the gridworld in Example \ref{e3}. For each trajectory, the environment begins in state one, and dynamically evolves under one of the policies for two hundred time periods. 
The parametric policy is optimized using CEM, and the optimal parameter values are found to be $r=(15,12,4,2,184)$; correctly identifying each state-control pair leading to the absorbing state at state sixteen. 

In Figure \ref{f5} (Left Panel), the parametric policy samples the transition probabilities of the transient states for $185$ time periods, avoiding the absorbing state during this time. During time period $186$, the environment moves to the absorbing state for the remaining $14$ time periods. The fifteen transient states each have four transition probabilities to learn (one for each of the four controls). We subtract two from this total, for the two transitions leading to the absorbing state, giving a total of 58 transition probabilities to learn over $185$ time periods. This corresponds to approximately $3.19$ samples per transition probability for the transient states, and $3.5$ for the absorbing state, on average. 

In Figure \ref{f5} (Center Panel), the rollout policy samples the transition probabilities of the transient states for slightly longer, at $187$ time periods. During time period $188$, the environment moves to the absorbing state for the remaining $12$ time periods. This corresponds to approximately $3.22$ samples per transition probability for the transient states, and $3$ for the absorbing state, on average. An optimal policy would divide $199$ time periods between $62$ transition probabilities, leading to approximately $3.21$ samples per transition probability for both transient and absorbing states, on average. Therefore, the rollout and parametric policies are close to optimal (but not optimal) in this case. 

In Figure \ref{f5} (Right Panel), the decrease in missing information is shown at each time period for both the parametric policy and the rollout policy. Over two hundred time periods the missing information decreases from $256$ to $24.6$ for the parametric policy, and from $256$ to $24.1$ for the rollout policy. The rollout policy only slightly improves the parametric policy since it is already near-optimal. 

\subsection{Example 4}

This example is given by a maze with three absorbing states, as shown in Example \ref{e4}. Each CMC transition probability is one for a move of one square along the cardinal direction specified by a control (i.e., `up', `down', `left', or `right'), and zero otherwise; unless that move would cross a wall of the maze, or the boundary; in which case, the agent will remain in the same square with probability one.
\begin{Example}
\centering
\begin{tikzpicture} [scale = 1.5]
\draw[step=0.5cm,color=gray] (0,0) grid (2.5,2.5);
\fill[red] (2,2) rectangle (2.5,2.5) {}; 
\fill[red] (0.5,1.5) rectangle (1,2) {}; 
\fill[red] (1.5,0.5) rectangle (2,1) {}; 
\draw[step=0.5cm,color=black,line width=2pt,cap=round,rounded corners=1pt] 
(0,0) -- (1,0)
(1.5,0) -- (2.5,0)
(2.5,0) -- (2.5,2.5)
(2.5,2.5) -- (1.5,2.5)
(1,2.5) -- (0,2.5)
(0,2.5) -- (0,0)
(1.5,2) -- (1.5,2.5)
(1.5,2) -- (2,2)
(0,2) -- (1,2)
(1,2) -- (1,1.5)
(1,1.5) -- (2,1.5)
(0.5,1) -- (0.5,1.5)
(0,1) -- (0.5,1)
(1,1) -- (1.5,1)
(1,0.5) -- (1,1)
(2,0.5) -- (2,1)
(0.5,0.5) -- (2,0.5)
(1.5,0) -- (1.5,0.5);
\end{tikzpicture}
\caption{A $5\times 5$ maze with three absorbing states (red squares). Each CMC transition probability is one for a move of one square along the cardinal direction specified by a control (i.e., `up', `down', `left', or `right'), and zero otherwise; unless that move would cross a wall of the maze, or the boundary; in which case, the agent will remain in the same square with probability one.}
\label{e4}
\end{Example}

Now assume we know nothing about the transition probabilities except that there are three unknown states (and three unknown controls) that lead to absorbing states with unknown locations. The parameters are then: $r=(r_1,...,r_7)$; where the first three parameters correspond to the three unknown states, the next three parameters correspond to one unknown control for each of these states, and the last parameter corresponds to a single unknown time constant. 

\begin{figure}
\centering
\includegraphics[scale=0.8,bb=0 320 580 520,clip=true]{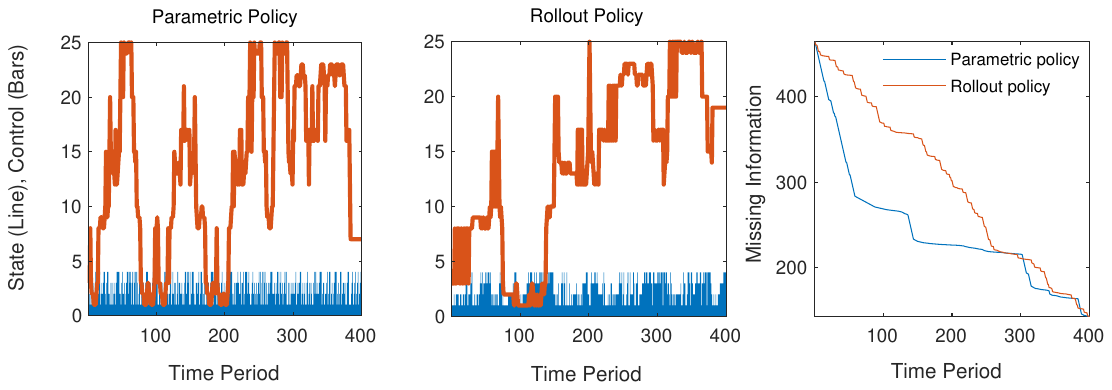}
\caption{Near-optimal policies for Example \ref{e4}. (Left Panel) Parametric policy applied to the CMC corresponding to the maze in Example \ref{e4}. (Center Panel) Rollout policy. (Right Panel) Decrease in missing information for each policy.}
\label{f6}
\end{figure}

Figure \ref{f6} shows two state trajectories for the CMC corresponding to the maze and absorbing states in Example \ref{e4}. For each trajectory, the environment begins in state three, and dynamically evolves under one of the policies for four hundred time periods. 
The parametric policy is optimized using CEM, and the optimal parameter values are found to be $r=(10,12,14,1,1,2,361)$; correctly identifying each state-control pair leading to an absorbing state at states five, seven, and ninteen. 

In Figure \ref{f6} (Left Panel), the parametric policy samples the transition probabilities of the transient states for $383$ time periods, avoiding each of the absorbing states during this time. During time period $384$, the environment moves to the absorbing state at state seven for the remaining $16$ time periods. The twenty two transient states each have four transition probabilities to learn. We subtract three from the total, for the three transitions leading to the absorbing states, giving a total of $85$ transition probabilities to learn over $383$ time periods. This corresponds to approximately $4.51$ samples per transition probability for the transient states, and $4$ for the absorbing state at state seven, on average. 

In Figure \ref{f6} (Center Panel), the rollout policy samples the transition probabilities of the transient states for slightly shorter, at $379$ time periods. During time period $380$, the environment moves to the absorbing state at state ninteen for the remaining $20$ time periods. This corresponds to approximately $4.46$ samples per transition probability for the transient states, and $5$ for the absorbing state at state ninteen, on average. An optimal policy would divide $399$ time periods between $89$ transition probabilities, leading to approximately $4.48$ samples per transition probability for all transient states, and one of the absorbing states, on average. Therefore, the rollout and parametric policies are close to optimal in this case.  

In Figure \ref{f6} (Right Panel), the decrease in missing information is shown at each time period for both the parametric policy and the rollout policy. Over four hundred time periods the missing information decreases from $464$ to $143$ for both parametric and rollout policies. 

We now attempt to solve the maze in Example \ref{e4} by making use of the agent's estimate of the CMC from applying the parametric policy as shown in Figure \ref{f6}. The maze exit path is itself given by the optimal policy of an associated shortest path problem. The Bellman equation for this shortest path problem can be written as:
\begin{equation}
J^*(i) = \underset{u\in U(i)}{\operatorname{min}}\ \Big [g(i)+\alpha\sum_{j} \widehat{p}_{ij}(u,\boldsymbol{F})J^*(j)\Big ],\ \ \ \forall i,\label{bell}
\end{equation}
where $\alpha$ is a discount factor, $\widehat{p}_{ij}(u,\boldsymbol{F})$ is the agent's estimate of the transition probabilities, and $g(i)$ is the cost for each state $i$; and is given by $g(i) = 1$ for each state except the maze exit at $i=23$, which has a cost of zero (i.e., $g(23) = 0$). As an infinite-horizon shortest-path problem, an optimal policy would select controls to get to the maze exit as quickly as possible and stay there. For this to work obviously requires a good estimate of the CMC in Example \ref{e4}. In order to avoid the absorbing states we make use of the parameterized control set $U(i)=U_0(i,r^*)$ at $k=0$.

Th Bellman equation can be solved using exact policy iteration. During each iteration, we evaluate the current policy by solving the following system of linear equations,
\begin{equation}
J_{\mu}(i) = g(i)+\alpha\sum_{j} \widehat{p}_{ij}(\mu(i),\boldsymbol{F})J_{\mu}(j),\ \ \ \forall i,\label{policyev}
\end{equation}
which can be written in matrix-vector notation as
$
MJ=g,
$
where $M_{ij}=\delta_{ij}-\alpha\widehat{p}_{ij}(\mu(i),\boldsymbol{F})$, $J_i=J_{\mu}(i)$, and $g_i=g(i)$. After solving (\ref{policyev}) for $J_{\mu}$ starting with a random policy, policy $\mu^k(i)$ can be ``improved" with the following step:
\begin{equation}
\mu^{k+1}(i) = \underset{\ u\in U(i)}{\operatorname{arg\ min}}\ \Big [g(i)+\alpha\sum_j \widehat{p}_{ij}(u,\boldsymbol{F})J_{\mu^k}(j)\Big ].
\end{equation}
This process of policy evaluation, followed by policy improvement, is repeated until the policy converges: at which stage it is the optimal policy $\mu^*(i)$ solving the Bellman equation.

In Figure \ref{f7}, the optimal policy solving the Bellman equation is shown as the blue line with arrows for $\alpha=0.99$. In this case, the optimal policy gives the unique shortest path starting at the maze entrance (state three) and finishing at the maze exit (state 23). 

In Figure \ref{f9}, the maze given by Example \ref{e4} has been modified slightly: the maze exit (state 23) now has a left wall instead of a right wall. Applying the original parametric policy to this modified CMC updates the CMC-estimate online, and causes the maze exit path to be re-routed by the optimal policy solving Bellman's equation (blue line with arrows in Figure \ref{f9}). This example demonstrates that parametric policies can be used online to dynamically update CMC estimates though informative exploration.  
\begin{figure}
\centering
\begin{tikzpicture} [scale = 1.5]
\draw[step=0.5cm,color=gray] (0,0) grid (2.5,2.5);
\fill[red] (2,2) rectangle (2.5,2.5) {}; 
\fill[red] (0.5,1.5) rectangle (1,2) {}; 
\fill[red] (1.5,0.5) rectangle (2,1) {}; 
\draw[step=0.5cm,color=black,line width=2pt,cap=round,rounded corners=1pt] 
(0,0) -- (1,0)
(1.5,0) -- (2.5,0)
(2.5,0) -- (2.5,2.5)
(2.5,2.5) -- (1.5,2.5)
(1,2.5) -- (0,2.5)
(0,2.5) -- (0,0)
(1.5,2) -- (1.5,2.5)
(1.5,2) -- (2,2)
(0,2) -- (1,2)
(1,2) -- (1,1.5)
(1,1.5) -- (2,1.5)
(0.5,1) -- (0.5,1.5)
(0,1) -- (0.5,1)
(1,1) -- (1.5,1)
(1,0.5) -- (1,1)
(2,0.5) -- (2,1)
(0.5,0.5) -- (2,0.5)
(1.5,0) -- (1.5,0.5);
\draw[->,line width=1pt,color=blue] (1.25,2.25) -- (1.25,2);
\draw[->,line width=1pt,color=blue] (2.25,1.75) -- (2.25,1.5);
\draw[->,line width=1pt,color=blue] (2.25,1.25) -- (1,1.25);
\draw[->,line width=1pt,color=blue] ((0.25,0.75) -- (0.25,0.5);
\draw[->,line width=1pt,color=blue] (0.25,0.25) -- (1,0.25);
\draw[step=0.5cm,color=blue,line width=1pt,cap=round,rounded corners=1pt]
(1.25,2.25) -- (1.25,1.75)
(1.25,1.75) -- (2.25,1.75)
(2.25,1.75) -- (2.25,1.25)
(2.25,1.25) -- (0.75,1.25)
(0.75,1.25) -- (0.75,0.75)
(0.75,0.75) -- (0.25,0.75)
(0.25,0.75) -- (0.25,0.25)
(0.25,0.25) -- (1.25,0.25)
(1.25,0.25) -- (1.25,0);
\end{tikzpicture}
\caption{Optimal policy for maze exit path (blue line with arrows) found after applying the parametric policy (shown in Figure \ref{f6}) to estimate the CMC in Example \ref{e4}.}
\label{f7}
\end{figure}

\begin{figure}
\centering
\begin{tikzpicture} [scale = 1.5]
\draw[step=0.5cm,color=gray] (0,0) grid (2.5,2.5);
\fill[red] (2,2) rectangle (2.5,2.5) {}; 
\fill[red] (0.5,1.5) rectangle (1,2) {}; 
\fill[red] (1.5,0.5) rectangle (2,1) {}; 
\draw[step=0.5cm,color=black,line width=2pt,cap=round,rounded corners=1pt] 
(0,0) -- (1,0)
(1.5,0) -- (2.5,0)
(2.5,0) -- (2.5,2.5)
(2.5,2.5) -- (1.5,2.5)
(1,2.5) -- (0,2.5)
(0,2.5) -- (0,0)
(1.5,2) -- (1.5,2.5)
(1.5,2) -- (2,2)
(0,2) -- (1,2)
(1,2) -- (1,1.5)
(1,1.5) -- (2,1.5)
(0.5,1) -- (0.5,1.5)
(0,1) -- (0.5,1)
(1,1) -- (1.5,1)
(1,0.5) -- (1,1)
(2,0.5) -- (2,1)
(0.5,0.5) -- (2,0.5)
(1,0) -- (1,0.5);
\draw[->,line width=1pt,color=blue] (1.25,2.25) -- (1.25,2);
\draw[->,line width=1pt,color=blue] (2.25,1.75) -- (2.25,1.5);
\draw[->,line width=1pt,color=blue] (2.25,0.75) -- (2.25,0.5);
\draw[->,line width=1pt,color=blue] (1.75,0.25) -- (1.5,0.25);
\draw[step=0.5cm,color=blue,line width=1pt,cap=round,rounded corners=1pt]
(1.25,2.25) -- (1.25,1.75)
(1.25,1.75) -- (2.25,1.75)
(2.25,1.75) -- (2.25,0.25)
(2.25,0.25) -- (1.25,0.25)
(1.25,0.25) -- (1.25,0);
\end{tikzpicture}
\caption{Dynamic Re-routing. Optimal policy leading to a re-routed maze exit path (blue line with arrows) found after applying the original parametric policy to a slightly modified CMC (see text for details).}
\label{f9}
\end{figure}

\subsection{Example 5}

This example is given by a maze similar to Example \ref{e4}, but with two \emph{non-backtracking states} placed along the maze exit path, as shown in Example \ref{e5}. The transition probabilities for a non-backtracking state do not allow an agent to revisit its previous state (i.e., it cannot backtrack). The non-backtracking states in Example \ref{e4} break up the maze into three distinct regions; so that after each region has been explored and the agent moves to the next region it is no longer possible to return to previous regions. Assuming an agent starts at the maze entrance, it can only exit a non-backtracking state along a single cardinal direction (given by a blue arrow) due to the placement of non-backtracking states and maze walls in this example. This type of irreversible behaviour is motivated by Example \ref{e2}. 
\begin{Example}
\centering
\begin{tikzpicture} [scale = 1.5]
\draw[step=0.5cm,color=gray] (0,0) grid (2.5,2.5);
\fill[green] (1,1) rectangle (1.5,1.5) {}; 
\fill[green] (2,1) rectangle (2.5,1.5) {}; 
\draw[step=0.5cm,color=black,line width=2pt,cap=round,rounded corners=1pt] 
(0,0) -- (1,0)
(1.5,0) -- (2.5,0)
(2.5,0) -- (2.5,2.5)
(2.5,2.5) -- (1.5,2.5)
(1,2.5) -- (0,2.5)
(0,2.5) -- (0,0)
(1.5,2) -- (1.5,2.5)
(1.5,2) -- (2,2)
(0,2) -- (1,2)
(1,2) -- (1,1.5)
(1,1.5) -- (2,1.5)
(0.5,1) -- (0.5,1.5)
(0,1) -- (0.5,1)
(1,1) -- (1.5,1)
(1,0.5) -- (1,1)
(2,1) -- (2.5,1)
(0.5,0.5) -- (2,0.5)
(1.5,0) -- (1.5,0.5);
\draw[->,line width=1pt,color=gray] (1.25,2.15) -- (1.25,2);
\draw[->,line width=1pt,color=gray] (2.25,1.65) -- (2.25,1.5);
\draw[->,line width=1pt,color=gray] ((0.25,0.65) -- (0.25,0.5);
\draw[->,line width=1pt,color=gray] (0.85,0.25) -- (1,0.25);
\draw[dashed,step=0.5cm,color=gray,line width=1pt,cap=round,rounded corners=1pt]
(1.25,2.25) -- (1.25,1.75)
(1.25,1.75) -- (2.25,1.75)
(2.25,1.75) -- (2.25,1.25)
(2.25,1.25) -- (0.75,1.25)
(0.75,1.25) -- (0.75,0.75)
(0.75,0.75) -- (0.25,0.75)
(0.25,0.75) -- (0.25,0.25)
(0.25,0.25) -- (1.25,0.25)
(1.25,0.25) -- (1.25,0);
\draw[->,line width=1.5pt,color=blue] (2.5,1.25) -- (2,1.25);
\draw[->,line width=1.5pt,color=blue] (1.5,1.25) -- (1,1.25);
\end{tikzpicture}
\caption{A $5\times 5$ maze with two non-backtracking states (green squares with blue arrows) along the maze exit path. Informative exploration is now broken into three distinct maze regions.}
\label{e5}
\end{Example}

Assuming we know nothing about the transition probabilities except that there are two unknown states (and two unknown controls), we now require six parameters: the first two correspond to unknown states, the next two correspond to an unknown control for each of these states, while the last two correspond to unknown time constants. 

\begin{figure}
\centering
\includegraphics[scale=0.8,bb=0 320 580 520,clip=true]{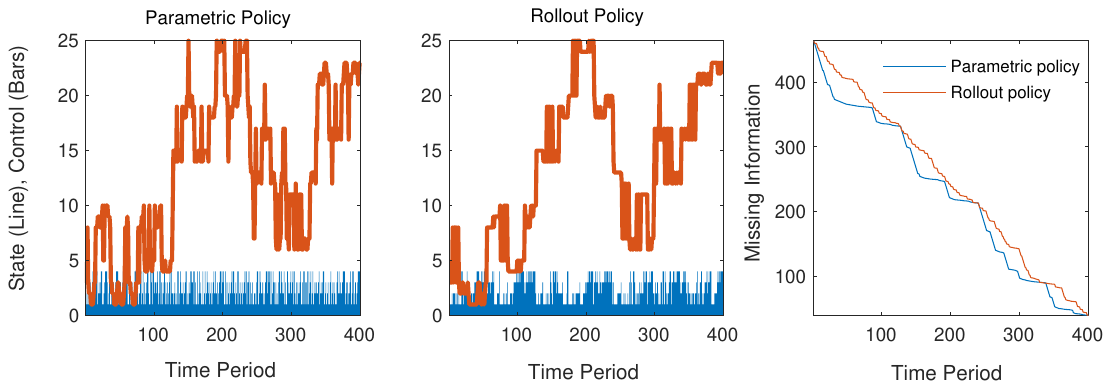}
\caption{Near-optimal policies for Example \ref{e5}. (Left Panel) Parametric policy applied to the CMC corresponding to the maze in Example \ref{e5}. (Center Panel) Rollout policy. (Right Panel) Decrease in missing information for each policy.}
\label{f8}
\end{figure}

Figure \ref{f8} shows two state trajectories for the CMC corresponding to the maze and non-backtracking states in Example \ref{e5}. For each trajectory, the environment begins in state three, and dynamically evolves under one of the policies for four hundred time periods. The parametric policy is optimized using CEM, and the optimal parameter values are found to be $r=(10,14,3,2,127,221)$; correctly identifying each state-control pair leading to a non-backtracking state at states fifteen and thirteen, respectively. 

In Figure \ref{f8} (Left and Middle Panels), the parametric and rollout policies sample the transition probabilities of the first maze region for $127$ time periods. During time period $127$, the environment moves from state ten to state fifteen, and both policies sample the transition probabilities of the second maze region for $113$ time periods. During time period $240$, the environment moves from state fourteen to state thirteen, and both policies sample the transition probabilities of the third maze region for $160$ time periods. An optimal policy would divide the $400$ time periods equally between the states in each maze region; leading to $128$ time periods for the first maze region, $112$ time periods for the second maze region, and $160$ for the third maze region. Therefore, the parametric and rollout policies are very close to optimal in this example. 

In Figure \ref{f6} (Right Panel), the decrease in missing information is shown at each time period for both the parametric policy and the rollout policy. Over four hundred time periods the missing information decreases from $464$ to $39.2$ for both parametric and rollout policies. 

\subsection{Example 6}

So far, we have investigated a diverse range of CMCs without giving much consideration to their possible applications. In our final example, we look at a simplified model for finding a car parking space. This example is taken from \cite{bert3}, and corresponds to the CMC shown in Example \ref{e6}. Model states represent whether a car parking space is either available $A$, or not available $\bar{A}$ at location $k$. If one is not available, the driver can continue to the next parking space at location $k+1$. If a parking space is available, the driver can either choose to park (thereby entering a terminal state $T$), or continue to location $k+1$ in the hope of finding a parking space closer to their destination (which is at location $k=N$ if we start from location $k=0$). However, if the driver has not found an available parking space by the time they reach location $k=N-1$, then they must pay the cost to park in an expensive car park. 
\begin{Example}
\centering
\begin{tikzpicture}[->,shorten >=2pt,line width=0.5pt,node distance=2cm]
\node [circle,draw] (one) {$A$};
\node [circle,draw] (two) [right of=one] {$\bar{A}$};
\path (one) edge [loop left] node [left] {$0.25$} (one);
\path (two) edge [bend right] node [above] {$0.25$} (one);
\path (one) edge [bend right] node [below] {$0.75$} (two);
\path (two) edge [loop right] node [right] {$0.75$} (two);
\node at (0,1.4) {\color{blue}{$u=\text{continue}$}};
\node [circle,draw] (two-u2) [below of=one] {$A$};
\node [circle,draw] (three-u2) [below of=two] {$T$};
\path (two-u2) edge [line to] node [above] {$1$} (three-u2);
\path (three-u2) edge [loop right] node [right] {$1$} (three-u2);
\node at (-0.2,-1.2) {\color{blue}{$u=\text{park}$}};
\end{tikzpicture}
\caption{CMC for finding a car park.}
\label{e6}
\end{Example}

Assume, for arguments sake, we know nothing about the transition probabilities for Example \ref{e6} except that one of the three states leads to an absorbing state under one of the two available controls. To learn the transition probabilities of this model by applying the parametric policy now requires three parameters: one for the unknown state leading to the absorbing state, one for the unknown control from that state, and one for the unknown time constant. 

Figure \ref{f9} shows a ``typical" state trajectory under each of the policies for the CMC in Example \ref{e6}. In each case, the environment begins in state $\bar{A}$ (state one) and stochastically evolves under one of the policies for eighty time periods. The parametric policy is optimized using exhaustive search with Monte Carlo averaging (over 1000 trajectories), and the optimal parameter values are found to be $r=(2,2,56)$. The parameters correctly identify state two ($A$) and control two (``park") as leading to the absorbing state given by state $T$. Since this problem is stochastic, the value of the time constant now depends both on the number of transition probabilities for each control, as well as their relative probability values. While this policy is likely to be optimal, this means we can no longer demonstrate it using simple counting arguments -- only that more time should be spent exploring the chain corresponding to control one (``continue") than the chain corresponding to control two (``park"). The learned time constant suggests $55$ time periods is sufficient for control one, and $25$ time periods for control two.
\begin{figure}
\centering
\includegraphics[scale=0.8,bb=0 320 580 520,clip=true]{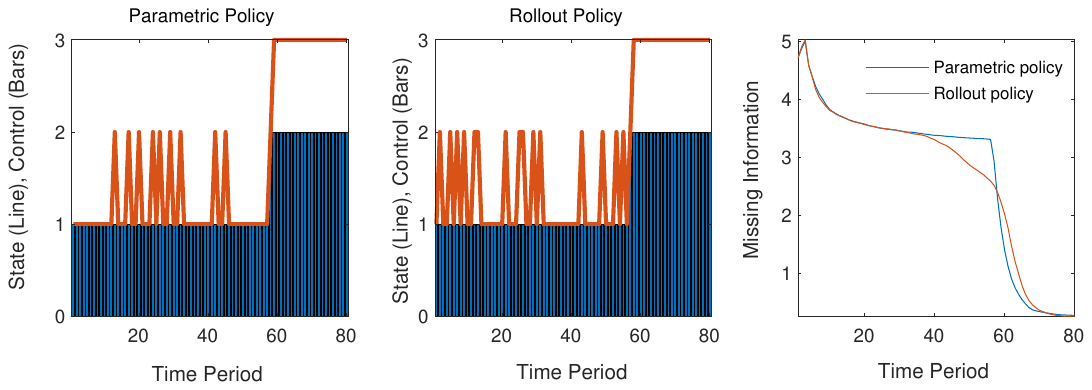}
\caption{Policies for Example \ref{e6}. (Left Panel) A typical trajectory under the parametric policy. (Center Panel) The corresponding rollout policy. (Right Panel) Decrease in Monte Carlo averaged missing information for each policy.}
\label{f9}
\end{figure}

In Figure \ref{f9} (Left Panel), a typical trajectory under the parametric policy is shown. When the environment is in state one the policy selects control one, allowing the agent to explore the transition probabilities of the chain corresponding to ``continue". In this case it is clear from the figure that the agent spends most of its time in state one, with an occasional stochastic transition to state two before returning to state one. Similar behaviour can be seen in Figure \ref{f9} (Center Panel) for the corresponding rollout policy. Control two is applied only when the environment is in state two and the time period is at least $56$ (which happens to be $58$ for the parametric policy, and $57$ for the rollout policy), causing the environment to move to state three for the remaining time. In Figure \ref{f9} (Right Panel), the decrease in missing information for each policy is shown after Monte Carlo averaging. Starting at $4.7$, the missing information decreases to $0.28$ for the parametric policy, and $0.27$ for the rollout policy.

\begin{figure}
\centering
\includegraphics[scale=0.8,bb=0 320 580 520,clip=true]{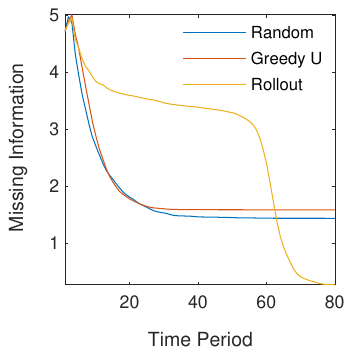}
\caption{Suboptimal policies for Example \ref{e6}. The decrease in the Monte Carlo averaged missing information is shown for the uniform random policy over unrestricted controls (Random), the greedy policy over unrestricted controls (Greedy U). The rollout policy with a Greedy U base policy is shown for comparison (Rollout).}
\label{f10}
\end{figure}

A comparison with two suboptimal policies is shown in Figure \ref{f10}. 
A uniform random policy over unrestricted controls is seen to marginally outperform greedy maximization of the information measure over unrestricted controls. However, neither policy performs very well and both lead to suboptimal exploration as they each prematurely choose ``park" before learning the transition probabilities for the chain corresponding to ``continue". The rollout policy for a base policy given by the greedy policy over unrestricted controls is also shown. It is clear from Figure \ref{f10} that the rollout policy is able to significantly improve its base policy in this case, generating a policy similar to that shown in Figure \ref{f9}. 

\subsection{Information Measures}

We intentionally kept the information measure $h(i,u,\boldsymbol{F})$ abstract in our work. Empirical evidence presented in \cite{little} suggests different forms for $h$ could be substituted into our equations and lead to qualitatively similar results. The information measure used in this work was the predicted information gain (PIG) \citep{little}. At each step, this measure compares two alternative CMC estimates: the current CMC estimate, and temporary updates of the current CMC estimate found from exploring new transitions. The larger the difference between these estimates, the greater the predicted information gain. Explicitly, PIG is given by
\begin{equation}
h(i,u,\boldsymbol{F})=\sum_{j^*} \widehat{p}_{ij^*}(u,\boldsymbol{F})\mathrm{D}_{\mathrm{KL}}[\widehat{p}_{i\cdot}(u,\boldsymbol{F}^{i\rightarrow j^*})||\widehat{p}_{i\cdot}(u,\boldsymbol{F}) ],
\label{pig}
\end{equation}
where $\mathrm{D}_{\mathrm{KL}}$ is the KL divergence \citep{mackay} between distributions $\widehat{p}_{i\cdot}(u,\boldsymbol{F})$ and $\widehat{p}_{i\cdot}(u,\boldsymbol{F}^{i\rightarrow j^*})$. The first distribution is the current estimate of $p_{i\cdot}(u)$ based on the counts in $\boldsymbol{F}$. The second distribution includes a temporary update of $F_{uij^*}$ to $F_{uij^*}+1$ due to exploring a transition from $i$ to $j^*$. If this transition has not been visited previously, the update means the two CMC estimates will now differ and the KL divergence will become greater than zero: thereby predicting an information gain from $p_{i\cdot}(u)$. 

The complexity of evaluating (\ref{pig}) is $O(s)$, where $s$ is the number of states $|S|$. For very large environments with many states this evaluation could become prohibitive. It would then be necessary to either use prior knowledge about the CMC, or introduce some form of approximation, to limit the size of the sum in (\ref{pig}). Alternatively, a less accurate information measure with more favourable complexity could be chosen.

\section{Conclusion and Discussion}
\label{conc}

We have given the general form of an optimal policy for an agent to learn controllable dynamics in an unknown environment by exploration. At each step, an agent would like to know where the best place is to explore next. Our suggestion is that exploring involves greedily maximizing an information measure by selecting controls from a constraint set that depends on time; i.e., some controls will only be available during certain times. The reason is due to the existence of states that limit or restrict control of the dynamics; such as transient states, absorbing states, and non-backtracking states. These states strongly influence exploration and need to be carefully taken into account. The proposed non-stationary policy is therefore essential for achieving optimal exploration over a limited time horizon. 

To support our proposal we investigated six examples of controllable dynamics in detail. By applying simple counting arguments, comparing with suboptimal policies, and making use of a sequential improvement property from dynamic programming, we were able to demonstrate optimality, or near-optimality, of our policy in most of these examples. 

The results presented in this work were derived using a particular optimization scheme given by the cross-entropy method. However, our results and conclusions are not expected to strongly depend on the use of this method and alternative choices should work equally well, or perhaps better. Precise optimization might not even be necessary in all cases. For example, when experimenting with very complex environments where only a few parameters were modelled (under-parameterized), or when parameter values were found imprecisely due to applying limited optimization resources (under-optimized), we often found ``good" suboptimal policies. This suggests the proposed policy might also be used more generally for reducing exploration in less favourable regions given limited resources for exploration. It also suggests parameterization and optimization need not be an all-or-nothing approach in order to be useful.

Our work emphasises the importance of a non-stationary policy for achieving optimality during exploration over a limited time horizon. By contrast, the reinforcement learning literature is generally concerned with stationary policies \citep{bert3}. This distinction is roughly due to the fact that in our case of resource-limited exploration we only have a finite time horizon to work over, and so non-stationary policies are natural to apply. While reinforcement learning tasks often involve (effectively) infinite horizons and therefore stationary policies become simpler to work with. For example, in the Q-Learning algorithm the optimal policy is derived from a Q-factor that does not depend explicitly on time. The reason is that to satisfy convergence requirements learning episodes are repeated many times over, generally with different initial conditions leading to different trajectories. The final Q-factor is therefore an average over many different trajectories; i.e., any distinct temporal properties of individual trajectories have been averaged out. On the other hand, by searching in policy space we work directly with individual trajectories, ultimately selecting the best one and retaining all of its unique properties. Several variants of Q-Learning achieve optimal exploration over an infinite horizon by using randomized policies (see, for example, \cite{storck1995reinforcement}). However, a randomized policy does not produce repeatable results due to its non-determinism, and individual trajectories are again averaged over. This behaviour is unsatisfactory for understanding why a particular policy should work well, or how best to explore an unknown environment over a finite horizon. The approach presented here specifically addresses these issues. 

This work should be of direct interest to researchers in robotics and automation, artificial intelligence, and mathematical modelling. It may also be relevant to researchers interested in exploration by animals or humans. 

\subsection*{Acknowledgements}

P.N.L would like to thank the Redwood Center for Theoretical Neuroscience for hosting a recent sabbatical visit.


\bibliographystyle{plainnat}
\bibliography{Loxley_references} 

\end{document}